\documentclass[final,5p,times,twocolumn]{elsarticle}

\usepackage{amsmath,amssymb,amsfonts}
\usepackage{mathtools}
\usepackage{graphicx}
\usepackage{booktabs}
\usepackage{multirow}
\usepackage{array}
\usepackage{hyperref}
\usepackage{caption}
\usepackage{subcaption}
\usepackage{geometry}
\usepackage{placeins}
\newcommand{\KL}{\mathrm{KL}}
\newcommand{\E}{\mathbb{E}}
\newcommand{\IG}{\mathrm{IG}}      
\newcommand{\NIG}{\mathrm{NIG}}    

\allowdisplaybreaks

\journal{Medical Image Analysis}

\begin{document}

\begin{frontmatter}

\title{ClinNet: Evidential Ordinal Regression with Bilateral Asymmetry and Prototype Memory for Knee Osteoarthritis Grading}

\author[inst1]{Xiaoyang Li\corref{cor1}}
\ead{[2666984@dundee.ac.uk]}
\author[inst1]{Runni Zhou}

\affiliation[inst1]{organization={College of Medicine and Biological Information Engineering, Northeastern University, Shenyang 110016, China.}}

\cortext[cor1]{Xiaoyang Li}

\begin{abstract}
Knee osteoarthritis (KOA) grading based on radiographic images is a critical yet challenging task due to subtle inter-grade differences, annotation uncertainty, and the inherently ordinal nature of disease progression.\cite{hunter2019osteoarthritis,cross2014global,kellgren1957radiological,gutierrez2016ordinalsurvey}
Conventional deep learning approaches typically formulate this problem as deterministic multi-class classification, ignoring both the continuous progression of degeneration and the uncertainty in expert annotations.\cite{tiulpin2018deepknee,antony2017quantifying,guo2017calibration,kendall2017uncertainties}
In this work, we propose ClinNet, a novel trustworthy framework that addresses KOA grading as an evidential ordinal regression problem.\cite{amini2020deepedlregression,sensoy2018edl,cao2020coral}
The proposed method integrates three key components: (1) a Bilateral Asymmetry Encoder (BAE) that explicitly models medial--lateral structural discrepancies;\cite{altman2007oarsi,woo2018cbam,hu2018senet}
(2) a Diagnostic Memory Bank that maintains class-wise prototypes to stabilize feature representations;\cite{snell2017prototypical,wu2018nonparametric,he2020moco}
and (3) an Evidential Ordinal Head based on the Normal-Inverse-Gamma (NIG) distribution to jointly estimate continuous KL grades and epistemic uncertainty.\cite{amini2020deepedlregression,kendall2017uncertainties}
Extensive experiments demonstrate that ClinNet achieves a Quadratic Weighted Kappa of 0.892 and Accuracy of 0.768, statistically outperforming state-of-the-art baselines ($p < 0.001$).\cite{cohen1968kappa,fleiss1973weightedkappa}
Crucially, we demonstrate that the model’s uncertainty estimates successfully flag out-of-distribution samples and potential misdiagnoses, paving the way for safe clinical deployment.\cite{hendrycks2017baselineood,ovadia2019can,geifman2017selective,elyaniv2010selective}
\end{abstract}


\end{frontmatter}

\section{Introduction}

\subsection{Clinical Background \& Motivation}
Knee osteoarthritis (KOA) is a chronic and progressive degenerative musculoskeletal disease and a leading cause of pain, mobility impairment, and functional disability among older adults worldwide.\cite{hunter2019osteoarthritis,cross2014global}
With the rapid aging of the global population, the prevalence of KOA continues to increase, placing a substantial burden on both patients and healthcare systems.\cite{cross2014global}
In routine clinical practice, radiography remains the most used imaging modality for KOA assessment due to its low cost and wide availability.\cite{hunter2019osteoarthritis,altman2007oarsi}
Among existing diagnostic criteria, the Kellgren--Lawrence (KL) grading system has been widely adopted as the gold standard for evaluating radiographic disease severity.\cite{kellgren1957radiological,altman2007oarsi}
Despite its widespread use, KL grading is inherently subjective, as it relies on visual assessment of multiple radiographic features, including joint space narrowing, osteophyte formation, subchondral sclerosis, and bony deformities.\cite{kellgren1957radiological,altman2007oarsi}
These features typically emerge gradually and heterogeneously, making precise grade assignment challenging.\cite{hunter2019osteoarthritis}
Prior studies have consistently reported notable inter-observer and intra-observer variability in KL grading, particularly in the early stages of the disease, where radiographic changes are subtle and poorly defined.\cite{landis1977measurement,wing2021reliability}
The ambiguity is most evident when differentiating KL grade 1 (doubtful KOA) from KL grade 2 (minimal KOA).\cite{kellgren1957radiological,wing2021reliability}
Conceptually, KL-1 is characterized by possible osteophytic lipping, whereas KL-2 requires definite osteophytes with possible joint space narrowing.\cite{kellgren1957radiological,altman2007oarsi}
In practice, however, the distinction between “possible” and “definite” osteophytes lacks a clear objective boundary.\cite{altman2007oarsi}
Early osteophytes are often small, low-contrast, and sensitive to imaging conditions such as projection angle and exposure, while mild joint space changes may be confounded by normal anatomical variability.\cite{altman2007oarsi,wing2021reliability}
Even experienced musculoskeletal radiologists frequently encounter difficulty in consistently distinguishing between these adjacent grades.\cite{wing2021reliability}
This uncertainty has important clinical implications, as the transition from KL-1 to KL-2 often marks the threshold for a radiographic diagnosis of KOA and may influence subsequent monitoring and intervention strategies.\cite{hunter2019osteoarthritis}

Radiographic assessment of KOA further relies heavily on intra-articular anatomical comparison within a single knee.\cite{altman2007oarsi}
In particular, the relative difference between the medial and lateral compartments plays a critical role in clinical interpretation.\cite{altman2007oarsi}
Due to asymmetric biomechanical loading, KOA most commonly affects the medial compartment first, leading to earlier joint space narrowing and osteophyte formation compared with the lateral compartment.\cite{hunter2019osteoarthritis,altman2007oarsi}
Consequently, radiologists frequently assess compartmental asymmetry to determine whether observed changes reflect pathological degeneration rather than normal variation, placing greater emphasis on relative structural differences than on absolute measurements alone.\cite{altman2007oarsi}
However, this clinically grounded reasoning is often insufficiently captured by existing automated KOA grading methods.\cite{tiulpin2020diagnostics,litjens2017survey}
Many deep learning approaches rely on cropped images of a single knee and treat the task as a generic image classification problem.\cite{tiulpin2018deepknee,antony2017quantifying,tiulpin2020diagnostics}
Such formulations tend to focus on localized appearance or global texture features, while failing to explicitly model structural relationships between the medial and lateral compartments.\cite{litjens2017survey}
This limitation is particularly detrimental for early-stage KOA, where absolute radiographic changes are minimal and subtle compartmental asymmetries provide crucial diagnostic cues.\cite{altman2007oarsi,wing2021reliability}
Moreover, the KL grading system represents an ordinal and progressive disease continuum, rather than a set of independent categorical labels.\cite{kellgren1957radiological,gutierrez2016ordinalsurvey}
Adjacent grades differ incrementally, yet many existing approaches formulate KOA grading as a standard multi-class classification task, disregarding the inherent ordinal structure.\cite{tiulpin2018deepknee}
This mismatch can lead to predictions that are inconsistent with clinical understanding of disease progression and limits the reliability of automated systems in real-world applications.\cite{niu2016ordinalcnn,cao2020coral}
Motivated by these clinical considerations, this study aims to develop an automated KOA assessment framework that better aligns with radiological practice by explicitly modeling the ordinal nature of KL grades and incorporating anatomically informed representations of medial–lateral compartmental structure.\cite{gutierrez2016ordinalsurvey,altman2007oarsi}
Such an approach seeks to improve early-stage discrimination, robustness, and clinical interpretability, thereby enhancing the translational potential of automated KOA grading systems.\cite{kendall2017uncertainties,ovadia2019can}

\subsection{Related Work}

\subsubsection{Traditional Radiographic Feature–Based KOA Assessment}
Early efforts in automated knee osteoarthritis (KOA) assessment were primarily based on handcrafted radiographic features.\cite{altman2007oarsi}
Commonly used descriptors included joint space width (JSW), osteophyte size and shape, subchondral bone density, edge irregularity, and texture-based statistics derived from predefined regions of interest.\cite{altman2007oarsi}
These features were typically combined with conventional machine learning models such as support vector machines, k-nearest neighbors, or random forests to perform KOA detection or severity grading.\cite{platt1999probabilistic}
While these approaches enabled quantitative analysis of radiographic characteristics, their performance was heavily dependent on the robustness of feature extraction pipelines.\cite{hunter2019osteoarthritis}
Variations in imaging acquisition, patient positioning, and anatomical morphology often led to unstable measurements.\cite{altman2007oarsi}
Moreover, handcrafted features generally capture isolated aspects of joint degeneration and fail to represent the complex, multifactorial progression of KOA.\cite{hunter2019osteoarthritis,cross2014global}
As a result, such methods demonstrated limited generalization ability and were gradually supplanted by data-driven deep learning approaches.\cite{litjens2017survey}

\subsubsection{CNN-Based Automated KOA Grading}
With the rapid advancement of deep learning, convolutional neural networks (CNNs) have become the dominant paradigm for automated KOA grading from knee radiographs.\cite{litjens2017survey,tiulpin2018deepknee,antony2017quantifying,tiulpin2020diagnostics}
Numerous studies have adopted backbone architectures such as ResNet, DenseNet, EfficientNet, and ConvNeXt to learn discriminative features in an end-to-end fashion.\cite{he2016resnet,huang2017densenet,tan2019efficientnet,liu2022convnext}
In most of these works, the KL grading task is formulated as a standard multi-class classification problem, where each KL grade is treated as an independent categorical label.\cite{tiulpin2018deepknee,antony2017quantifying}
CNN-based methods have achieved substantial performance improvements over traditional feature-based approaches on both public and private datasets, demonstrating strong representational capacity for radiographic patterns.\cite{litjens2017survey}
However, this formulation introduces several fundamental limitations. First, treating KL grades as mutually exclusive classes ignores the inherent ordinal nature of disease severity.\cite{gutierrez2016ordinalsurvey}
Adjacent grades represent incremental pathological progression, yet standard classification objectives fail to penalize inconsistent predictions that violate this ordering.\cite{niu2016ordinalcnn,cao2020coral}
Second, CNN classifiers typically rely on SoftMax outputs, which provide point estimates without explicitly modeling predictive uncertainty.\cite{guo2017calibration}
Consequently, models may produce highly confident yet incorrect predictions, a phenomenon that poses significant risks in clinical decision-making.\cite{kendall2017uncertainties,guo2017calibration}

\subsubsection{Ordinal Learning for Medical Severity Assessment}
To address the ordered structure of medical grading tasks, ordinal learning methods have been increasingly explored in medical image analysis.\cite{gutierrez2016ordinalsurvey}
Ordinal classification and ordinal regression frameworks explicitly incorporate label ordering constraints, often through cumulative link models, threshold-based formulations, or rank-consistent loss functions.\cite{niu2016ordinalcnn,cao2020coral}
Such approaches have been successfully applied to a range of medical applications, including tumor staging, disease severity scoring, and diabetic retinopathy grading.\cite{gutierrez2016ordinalsurvey}
Despite their demonstrated benefits, the adoption of ordinal learning in KOA grading remains relatively limited.\cite{tiulpin2020diagnostics}
Existing studies often focus on modifying loss functions or output encodings while still relying on conventional CNN backbones.\cite{niu2016ordinalcnn,cao2020coral}
Moreover, many ordinal methods do not account for the intrinsic ambiguity of early-stage KOA, where radiographic features overlap substantially across adjacent grades.\cite{wing2021reliability}
As a result, ordinal consistency alone is insufficient to fully address uncertainty and borderline cases commonly encountered in clinical practice.\cite{kendall2017uncertainties}

\subsubsection{Uncertainty Modeling and Trustworthy Medical AI}
In recent years, uncertainty estimation has emerged as a critical component of trustworthy medical artificial intelligence.\cite{kendall2017uncertainties,litjens2017survey}
Bayesian neural networks, Monte Carlo Dropout, ensemble learning, and evidential deep learning have been proposed to quantify predictive uncertainty, particularly in scenarios involving ambiguous inputs or distributional shifts.\cite{gal2016dropout,lakshminarayanan2017ensembles,sensoy2018edl,ovadia2019can}
These methods have shown value in medical imaging tasks such as radiological diagnosis, pathology grading, and lesion detection by reducing the risk of overconfident errors.\cite{kendall2017uncertainties}
Among these approaches, evidential learning has gained increasing attention due to its ability to model both prediction and uncertainty within a single deterministic network.\cite{sensoy2018edl}
By learning parameters of a higher-order probability distribution, evidential models can distinguish between confident and uncertain predictions without requiring repeated sampling.\cite{sensoy2018edl}
However, in the context of KOA grading, uncertainty-aware methods remain underexplored.\cite{tiulpin2020diagnostics}
Most existing approaches do not integrate uncertainty estimation with ordinal disease modeling, nor do they explicitly address the clinically challenging transition between early KL grades, such as KL-1 and KL-2.\cite{wing2021reliability,amini2020deepedlregression}

\subsubsection{Summary and Research Gap}
In summary, substantial progress has been made in automated KOA assessment through the adoption of deep learning techniques.\cite{litjens2017survey,tiulpin2018deepknee}
Nevertheless, several critical limitations persist. Traditional feature-based methods lack expressive power and robustness, while CNN-based classifiers overlook the ordinal structure of KL grades and fail to provide reliable uncertainty estimates.\cite{gutierrez2016ordinalsurvey,guo2017calibration,kendall2017uncertainties}
Although ordinal learning and uncertainty modeling have independently advanced in medical image analysis, their integration within KOA grading remains limited.\cite{cao2020coral,amini2020deepedlregression}
These shortcomings are particularly evident in early and borderline cases, where radiographic changes are subtle and inter-grade ambiguity is high.\cite{wing2021reliability}
Addressing these challenges requires a unified framework that simultaneously respects the ordinal nature of disease progression and provides principled uncertainty estimation, thereby aligning automated predictions more closely with clinical reasoning and decision-making requirements.\cite{kendall2017uncertainties,sensoy2018edl,amini2020deepedlregression}

\subsection{Our Contribution}
We propose ClinNet, a framework for knee OA KL grading that is designed to be both accurate and trustworthy, by explicitly leveraging the  ordinal nature  of KL labels and quantifying predictive uncertainty.\cite{gutierrez2016ordinalsurvey,kendall2017uncertainties}
\begin{itemize}
    \item \textbf{Evidential Ordinal Regression for KL grading.} We formulate KL grading as an evidential ordinal regression problem rather than a plain multi-class classification task.\cite{gutierrez2016ordinalsurvey,cao2020coral,amini2020deepedlregression}
    ClinNet predicts the parameters of a Normal–Inverse-Gamma (NIG) evidential distribution, enabling the model to output not only a continuous severity estimate (e.g., the regression mean) but also an evidence-driven measure of uncertainty.\cite{amini2020deepedlregression,sensoy2018edl}
    This supports more reliable decision-making, especially near class boundaries.\cite{kendall2017uncertainties}
    \item \textbf{Bilateral Asymmetry Encoder mimicking radiological comparison.} Inspired by clinical reading habits, we introduce a Bilateral Asymmetry Encoder (BAE) that learns to attend to two complementary regions (medial/lateral) and explicitly models their difference through a fused representation.\cite{altman2007oarsi,woo2018cbam,hu2018senet}
    This module emulates the radiologist’s comparative reasoning by extracting side-specific descriptors and an asymmetry cue, improving sensitivity to structural changes relevant to OA severity.\cite{altman2007oarsi}
    \item \textbf{Safety validation via uncertainty quantification.} To support safe deployment, we validate ClinNet’s trustworthiness by analyzing its uncertainty estimates derived from evidential parameters.\cite{kendall2017uncertainties,ovadia2019can}
    We demonstrate that uncertainty behaves as a meaningful safety signal—highlighting ambiguous or out-of-distribution cases—and can be used to support risk-aware usage, such as deferring low-confidence predictions for expert review.\cite{hendrycks2017baselineood,geifman2017selective,elyaniv2010selective}
\end{itemize}

\section{Methodology}

\subsection{Network Architecture}
ClinNet consists of a backbone, a BAE module, a Memory Bank, and an EDL Head.\cite{liu2022convnext,woo2018cbam,hu2018senet,snell2017prototypical,he2020moco,amini2020deepedlregression}

\begin{figure*}[t]
\centering
\includegraphics[width=\textwidth]{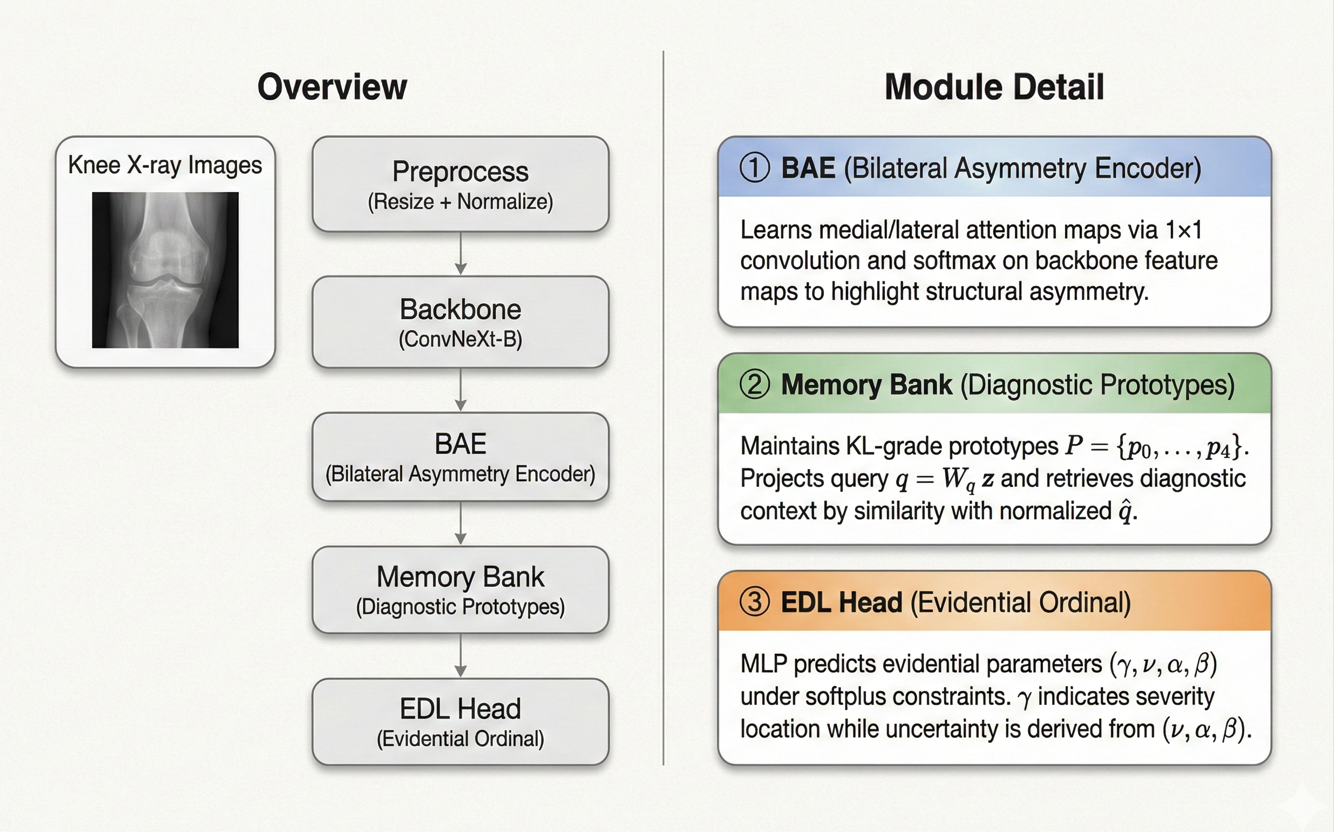}
\caption{Overview of the ClinNet architecture. The model takes paired knee images, extracts features via the BAE, aligns them using the Memory Bank, and outputs evidential parameters $\gamma,\nu,\alpha,\beta$.}
\label{fig:architecture}
\end{figure*}

\subsection{Bilateral Asymmetry Encoder (BAE)}
Given a knee radiograph $I$, the backbone network $B(\cdot)$ extracts a deep spatial feature map:
\begin{equation}
F=B(I) \in \mathbb{R}^{B \times C \times H \times W},
\end{equation}
where $B$ denotes the batch size, $C$ is the number of channels, and $H \times W$ is the spatial resolution.
The Bilateral Asymmetry Encoder (BAE) aims to mimic radiologists’ compartment-wise comparison by learning two attention branches that focus on the medial and lateral compartments, respectively, and by explicitly encoding their discrepancy.\cite{altman2007oarsi,woo2018cbam,hu2018senet}

(1) Medial/Lateral attention logits. We generate two attention logit maps using two learnable $1 \times 1$ convolutions:
\begin{equation}
\begin{aligned}
S_M &= g_M(F)=W_M*F, \\
S_{Lat} &= g_{Lat}(F)=W_{Lat}*F,
\end{aligned}
\end{equation}
where $W_M,W_{Lat} \in \mathbb{R}^{1 \times C \times 1 \times 1}$, and $S_M, S_{Lat} \in \mathbb{R}^{B \times 1 \times H \times W}$. Here, $*$ denotes convolution.

(2) Spatial softmax normalization. To obtain valid spatial attention distributions, we apply a softmax over all spatial locations.
Let $\text{vec}(\cdot)$ flatten the $H \times W$ grid into a vector of length $HW$. Then:
\begin{equation}
\begin{aligned}
\alpha_M &= \text{Softmax}(\text{vec}(S_M)) \in \mathbb{R}^{B \times HW}, \\
\alpha_{Lat} &= \text{Softmax}(\text{vec}(S_{Lat})) \in \mathbb{R}^{B \times HW}.
\end{aligned}
\end{equation}
After reshaping back to $\mathbb{R}^{B \times 1 \times H \times W}$, the attention maps satisfy:
\begin{equation}
\sum_{u=1}^{H}\sum_{v=1}^{W}\alpha_M(u,v)=1, \quad \sum_{u=1}^{H}\sum_{v=1}^{W}\alpha_{Lat}(u,v)=1.
\end{equation}
This “mass-conserving” design encourages each branch to concentrate on a compact set of discriminative locations.\cite{woo2018cbam}

(3) Attention pooling for compartment descriptors. Given $\alpha_M$ and $\alpha_{Lat}$, we compute two compartment-specific descriptors by weighted global aggregation:
\begin{equation}
\begin{aligned}
z_M &= \sum_{u=1}^{H}\sum_{v=1}^{W}\alpha_M(u,v) F(:,u,v) \in \mathbb{R}^{B \times C}, \\
z_{Lat} &= \sum_{u=1}^{H}\sum_{v=1}^{W}\alpha_{Lat}(u,v) F(:,u,v) \in \mathbb{R}^{B \times C},
\end{aligned}
\end{equation}
where $F(:,u,v)$ denotes the $C$-dimensional feature vector at location $uv$.
In practice, this is implemented as element-wise multiplication followed by summation over spatial dimensions.\cite{hu2018senet,woo2018cbam}

(4) Asymmetry feature. To explicitly encode medial–lateral discrepancy, we define the asymmetry feature as:
\begin{equation}
F_{asym}=|\text{Attention}_M(F)-\text{Attention}_{Lat}(F)|=|z_M-z_{Lat}| \in \mathbb{R}^{B \times C}.
\end{equation}
This term captures compartment-level imbalance that is clinically relevant for osteoarthritis grading (e.g., preferential medial joint-space narrowing).\cite{altman2007oarsi,hunter2019osteoarthritis}

(5) Fusion into a unified embedding. Finally, BAE concatenates the medial descriptor, lateral descriptor, and asymmetry feature and projects them into a unified representation:
\begin{equation}
h=\phi([z_M; z_{Lat}; F_{asym}]) \in \mathbb{R}^{B \times C},
\end{equation}
where $[ \cdot ; \cdot ]$ denotes concatenation and $\phi(\cdot)$ is an MLP with LayerNorm and GELU (optionally followed by dropout).
The BAE outputs $h$ as the compartment-aware embedding, and also provides $\alpha_M$ and $\alpha_{Lat}$ as interpretable attention maps.\cite{selvaraju2017gradcam}

\subsection{Evidential Ordinal Head (The Core)}
Modeling assumption. We treat the KL grade $y \in \mathbb{R}$ as an ordinal severity score and model it as a sample from a Gaussian likelihood with unknown mean and variance:\cite{gutierrez2016ordinalsurvey,amini2020deepedlregression}
\begin{equation}
y \mid \mu,\sigma^2 \sim \mathcal{N}(\mu,\sigma^2).
\end{equation}
To capture both predictive mean and uncertainty without sampling-based Bayesian inference, we place a Normal–Inverse-Gamma (NIG) prior over $\mu, \sigma^2$:\cite{amini2020deepedlregression,sensoy2018edl}
\begin{equation}
\mu \mid \sigma^2 \sim \mathcal{N}(\gamma, \sigma^2 / \nu), \quad \sigma^2 \sim \text{Inv-Gamma}(\alpha, \beta),
\end{equation}
where $\gamma \in \mathbb{R}$ is the location parameter, $\nu > 0$ controls the confidence on $\mu$, and $\alpha > 1, \beta > 0$ control the distribution over $\sigma^2$.
The evidential head predicts $\gamma, \nu, \alpha, \beta$ deterministically from features; in implementation, positivity constraints are enforced using softplus transforms and shifts (e.g., $\nu=\text{softplus}(\cdot)+\epsilon$, $\alpha=\text{softplus}(\cdot)+1+\epsilon$, $\beta=\text{softplus}(\cdot)+\epsilon$).\cite{amini2020deepedlregression}
This is standard in deep evidential regression.\cite{amini2020deepedlregression}

(1) NIG Marginal Likelihood and NLL Loss.
Because the NIG prior is conjugate to the Gaussian likelihood, we can integrate out $\mu$ and $\sigma^2$ and obtain the marginal likelihood of $y$ under the predicted evidential parameters:\cite{amini2020deepedlregression}
\begin{equation}
p(y \mid \gamma, \nu, \alpha, \beta) = \iint p(y \mid \mu, \sigma^2) p(\mu, \sigma^2 \mid \gamma, \nu, \alpha, \beta) \, d\mu \, d\sigma^2.
\end{equation}
This marginal corresponds to a Student-t predictive distribution (a heavy-tailed form that is robust to noise/outliers).\cite{amini2020deepedlregression}
For training, we minimize the negative log-likelihood (NLL):
\begin{equation}
\mathcal{L}_{NLL}(y; \gamma, \nu, \alpha, \beta) = -\log p(y \mid \gamma, \nu, \alpha, \beta).
\end{equation}
A commonly used closed-form NLL (equivalent to the Student-t marginal under the NIG parameterization above) is:
\begin{equation}
\begin{split}
\mathcal{L}_{NLL} = &\frac{1}{2}\log\left(\frac{\pi}{\nu}\right) - \alpha\log(2\beta(1+\nu)) \\
&+ \left(\alpha + \frac{1}{2}\right)\log(\nu(y-\gamma)^2 + 2\beta(1+\nu)) \\
&+ \log\Gamma(\alpha) - \log\Gamma\left(\alpha + \frac{1}{2}\right).
\end{split}
\end{equation}
This expression matches the NIG-based evidential regression objective widely adopted in the evidential learning literature.\cite{amini2020deepedlregression}

Interpretation.
$\gamma$ acts as the predicted severity score (later discretized to $0, \dots, 4$ if needed).\cite{gutierrez2016ordinalsurvey}
$\nu, \alpha, \beta$ Encode evidence: higher evidence typically yields sharper predictive distributions (lower uncertainty), while low evidence results in higher uncertainty.\cite{amini2020deepedlregression,kendall2017uncertainties}
This is the core mechanism that makes the head “trustworthy”: the network can express “I do not know” by producing low evidence.\cite{sensoy2018edl,kendall2017uncertainties}

(2) KL-Divergence Regularization Toward a Uniform (Low-Evidence) Prior.
Pure NLL minimization may allow the network to become overconfident.\cite{guo2017calibration}
To discourage unwarranted evidence, we add a KL regularization that pulls the predicted NIG toward a weak, near-uniform (i.e., low-evidence) reference prior:\cite{amini2020deepedlregression}
\begin{equation}
\mathcal{L}_{KL} = \text{KL}[ \text{NIG}(\gamma, \nu, \alpha, \beta) \parallel \text{NIG}_{\text{uniform}}(\gamma_0, \nu_0, \alpha_0, \beta_0) ].
\end{equation}
Here, $\text{NIG}_{\text{uniform}}$ is chosen to represent minimal prior commitment (e.g., small $\nu_0$, $\alpha_0$ close to 1, and a broad scale $\beta_0$), so that the model is penalized when it assigns strong evidence without sufficient support.\cite{amini2020deepedlregression}
A convenient way to compute KL uses the factorization:
\begin{equation}
p(\mu, \sigma^2) = p(\mu \mid \sigma^2) p(\sigma^2),
\end{equation}
which yields
\begin{equation}
\begin{aligned}
\KL[\NIG \,\|\, \NIG_0]
&= \KL[\IG(\alpha,\beta)\,\|\,\IG(\alpha_0,\beta_0)] \\
&\quad + \E_{\sigma^2 \sim \IG(\alpha,\beta)}
\Bigl[
  \KL\!\left[\mathcal{N}(\gamma,\sigma^2/\nu)\,\|\,\mathcal{N}(\gamma_0,\sigma^2/\nu_0)\right]
\Bigr].
\end{aligned}
\end{equation}

For the conditional Gaussian term, using the standard KL between univariate Gaussians and taking expectation over $\sigma^2$ under the inverse-gamma yields:
\begin{equation}
\begin{split}
\mathbb{E}[\text{KL}(\mathcal{N} \parallel \mathcal{N}_0)] &= \frac{1}{2}\left(\frac{\nu_0}{\nu} - \log\frac{\nu_0}{\nu} - 1\right) + \frac{1}{2}\nu_0(\gamma-\gamma_0)^2 \mathbb{E}\left[\frac{1}{\sigma^2}\right], \\
\mathbb{E}\left[\frac{1}{\sigma^2}\right] &= \frac{\alpha}{\beta}.
\end{split}
\end{equation}
Thus,
\begin{equation}
\mathbb{E}[\text{KL}(\mathcal{N} \parallel \mathcal{N}_0)] = \frac{1}{2}\left(\frac{\nu_0}{\nu} - \log\frac{\nu_0}{\nu} - 1\right) + \frac{1}{2}\nu_0(\gamma-\gamma_0)^2 \frac{\alpha}{\beta}.
\end{equation}
For the inverse-gamma term, one may exploit the bijection $t=1/\sigma^2$, under which $\sigma^2 \sim \text{Inv-Gamma}(\alpha, \beta)$ corresponds to $t \sim \text{Gamma}(\alpha, \beta)$ (shape $\alpha$, rate $\beta$).
Since KL is invariant under bijective transforms, we can compute:
\begin{equation}
\begin{aligned}
\KL[\IG(\alpha,\beta)\,\|\,\IG(\alpha_0,\beta_0)]
&= \KL[\Gamma(\alpha,\beta)\,\|\,\Gamma(\alpha_0,\beta_0)] .
\end{aligned}
\end{equation}

and use the closed-form KL for gamma distributions:
\begin{equation}
\text{KL}_{IG} = \alpha_0 \ln\frac{\beta}{\beta_0} - \ln\frac{\Gamma(\alpha)}{\Gamma(\alpha_0)} + (\alpha - \alpha_0)\psi(\alpha) - (\beta - \beta_0)\frac{\alpha}{\beta},
\end{equation}
where $\psi(\cdot)$ is the digamma function.
Putting the pieces together gives a computable $\mathcal{L}_{KL}$. The overall evidential loss is then:
\begin{equation}
\mathcal{L} = \mathcal{L}_{NLL} + \lambda_{KL} \mathcal{L}_{KL},
\end{equation}
where $\lambda_{KL}$ controls the strength of the “return-to-ignorance” constraint.
Practical note (alignment with implementation). In practice, many evidential regression implementations replace explicit $\mathcal{L}_{KL}$ with an evidence penalty proportional to prediction error (encouraging the network to output low evidence when it is wrong), which has similar intent and is computationally simpler.\cite{amini2020deepedlregression}

\begin{figure*}[t]
  \centering
  \includegraphics[width=\textwidth]{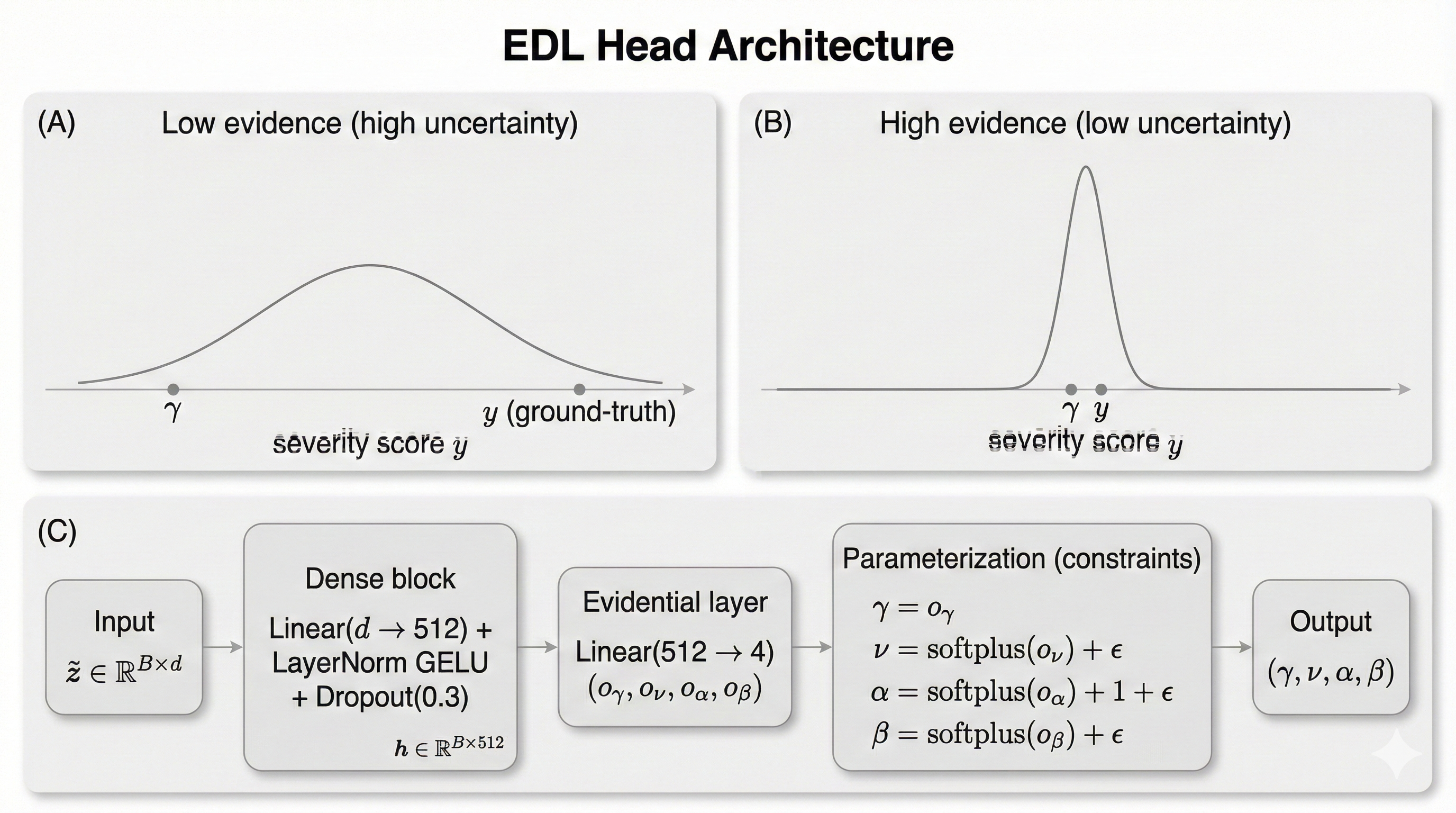}
  \caption{The Evidential Learning process. Unlike Softmax, the evidential (EDL) head predicts a distribution over possible severity scores, producing $\gamma,\nu,\alpha,\beta$ and enabling uncertainty estimation through evidence strength.\cite{sensoy2018edl,amini2020deepedlregression}}
  \label{fig:edl}
\end{figure*}

\section{Experiments and Results}

\subsection{Experimental Setup}
We conducted all experiments on the [Dataset] knee X-ray dataset, following a fixed train/validation/test split to ensure a fair and reproducible comparison across methods.\cite{oai2008protocol}
Each image was resized to 448×448 and normalized using ImageNet statistics.\cite{deng2009imagenet}
During training, we applied standard radiographic augmentation, including random horizontal flipping, mild affine transformation (rotation/translation/scale), brightness/contrast jittering, and random erasing, which improves generalization under acquisition variability.\cite{hendrycks2019imagenetc}
Validation and testing used only deterministic resizing and normalization.
ClinNet adopts a ConvNeXt-Base backbone (initialized from local pretrained weights) and outputs an evidential ordinal prediction through three key components: the Bilateral Asymmetry Encoder (BAE), a Diagnostic Memory Bank, and an Evidential Ordinal Head based on a Normal-Inverse-Gamma (NIG) parameterization.\cite{liu2022convnext,amini2020deepedlregression}
The model was trained for 300 epochs using AdamW with weight decay $1e-3$ and an initial learning rate $1e-5$, where the backbone learning rate was set to $0.1 \times$ the head modules.\cite{loshchilov2019adamw,kingma2015adam}
We used cosine annealing warm restarts for scheduling.\cite{loshchilov2017sgdr}
Mixed-precision training and gradient clipping were enabled for stability, and all experiments were run with a fixed random seed for reproducibility.
To respect the clinical evaluation protocol, the test split was strictly isolated during training: ClinNet selection and checkpointing were performed only according to the validation quadratic weighted kappa (QWK), and the test set was used only once for final reporting.\cite{cohen1968kappa,fleiss1973weightedkappa}
We evaluated ClinNet under both ordinal multi-class grading (KL 0–4) and a clinically meaningful binary OA detection task (OA defined as KL $\ge$ 2).\cite{hunter2019osteoarthritis}
For multi-class grading, we report Accuracy, QWK, Macro-F1, and MSE of the continuous severity estimate.\cite{cohen1968kappa,fleiss1973weightedkappa}
For binary OA detection, we report AUROC.\cite{hanley1982auc,delong1988roc}
In addition, we include reliability and safety analyses using epistemic uncertainty derived from the evidential head, together with robustness tests under distribution shift (e.g., rotation-based OOD) and image corruption (Gaussian noise/blur), which collectively probe the trustworthiness of the model beyond raw accuracy.\cite{guo2017calibration,brier1950verification,hendrycks2017baselineood,liang2018odin,lee2018mahalanobis,ovadia2019can}
Baseline models include representative CNN and Transformer architectures, namely DenseNet121, ResNet-50, InceptionV3, VGG16, Swin-T, ViT, ConvNeXt, and EfficientNet-V3, trained and evaluated under the same data split and metric definitions.\cite{huang2017densenet,he2016resnet,szegedy2016inception,simonyan2015vgg,liu2021swin,dosovitskiy2021vit,liu2022convnext,tan2019efficientnet}
This design ensures that performance gains can be attributed to the proposed ClinNet components rather than differences in data access or evaluation procedures.\cite{litjens2017survey}

\subsection{Comparative Performance with SOTA}

\begin{table}[t]
\centering
\caption{Comparison of diagnostic metrics. ACC is reported as percentage, and Kappa corresponds to quadratic weighted kappa (QWK).\cite{cohen1968kappa,fleiss1973weightedkappa}}
\label{tab:results}
\begin{tabular}{lcccc}
\toprule
\textbf{Method} & \textbf{ACC(\%)} & \textbf{Kappa} & \textbf{F1-score} & \textbf{Recall} \\
\midrule
ClinNet (Ours) & \textbf{76.897} & \textbf{0.892} & \textbf{0.781} & \textbf{0.769} \\
EfficientNet-V3 & 71.970 & 0.849 & 0.697 & 0.720 \\
InceptionV3 & 69.330 & 0.839 & 0.680 & 0.693 \\
Swin-T & 70.490 & 0.834 & 0.677 & 0.705 \\
VGG16 & 69.860 & 0.831 & 0.681 & 0.699 \\
ConvNeXt & 70.050 & 0.830 & 0.678 & 0.701 \\
DenseNet121 & 69.010 & 0.822 & 0.665 & 0.690 \\
ViT & 68.790 & 0.816 & 0.656 & 0.688 \\
ResNet50 & 63.640 & 0.757 & 0.593 & 0.636 \\
\bottomrule
\end{tabular}
\end{table}

We first benchmark ClinNet against a broad set of competitive CNN and Transformer baselines, including DenseNet121, ResNet-50, InceptionV3, VGG16, Swin-T, ViT, ConvNeXt, and EfficientNet-V3.\cite{huang2017densenet,he2016resnet,szegedy2016inception,simonyan2015vgg,liu2021swin,dosovitskiy2021vit,liu2022convnext,tan2019efficientnet}
All models were evaluated using the same split and metric definitions.
Table 1 summarizes multi-class diagnostic performance under KL grading, where ClinNet achieves the highest overall accuracy and agreement.\cite{cohen1968kappa,fleiss1973weightedkappa}
In particular, ClinNet reaches an Accuracy of 76.90\% and a QWK of 0.892, indicating substantially improved ordinal consistency compared to strong baselines such as EfficientNet-V3 (Accuracy 71.97\%, Kappa 0.8485) and Swin-T (Accuracy 70.49\%, Kappa 0.834).\cite{gutierrez2016ordinalsurvey,cao2020coral}
ClinNet also yields the best macro-F1 (0.781), reflecting balanced improvements across all severity grades rather than gains driven by majority classes.\cite{litjens2017survey}
Beyond single-number metrics, we evaluate “holistic superiority” by jointly comparing multi-dimensional capability profiles.
Specifically, we construct a radar chart that includes Accuracy, Kappa (QWK), F1-score, and Recall.
ClinNet consistently dominates the baselines across all axes, suggesting that its performance gains are not a metric-specific artifact but arise from a more robust and clinically aligned representation of ordinal severity.\cite{gutierrez2016ordinalsurvey}
This result supports the central hypothesis of ClinNet: incorporating compartment-aware asymmetry modeling and prototype-based memory alignment improves the reliability of radiographic grading.\cite{altman2007oarsi,snell2017prototypical,wu2018nonparametric}

\begin{figure}[t]
\centering
\includegraphics[width=\linewidth]{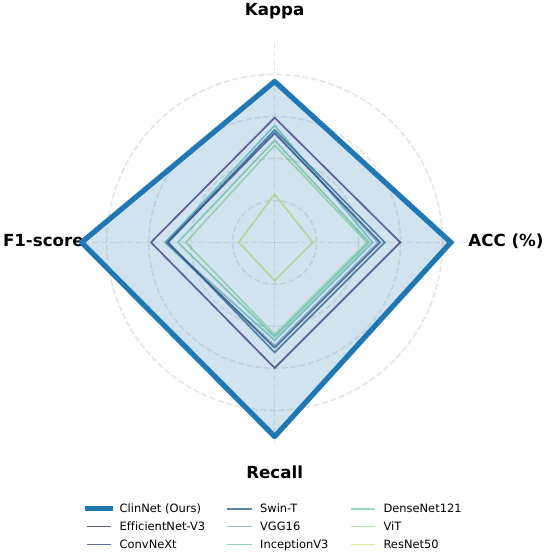}
\caption{Radar Chart Comparison (Normalized). ClinNet demonstrates superior performance across Accuracy, Kappa, F1, and Recall.}
\label{fig:radar}
\end{figure}

\subsection{Detailed Diagnostic Analysis}
Although overall metrics summarize performance, clinical deployment requires that the model behaves sensibly on ambiguous and borderline cases.\cite{wing2021reliability,kendall2017uncertainties}
We therefore conducted a detailed diagnostic analysis from complementary perspectives.
First, we evaluated discriminative ability on a clinically meaningful binary endpoint—OA detection (KL \(\ge\) 2)—using bootstrapped ROC curves.\cite{hanley1982auc,delong1988roc}
This analysis reports not only the mean ROC but also the 95\% confidence band, which reflects the stability of sensitivity–specificity trade-offs under sampling variability.\cite{delong1988roc}
Second, we further assessed performance under class-imbalance–sensitive conditions using the precision–recall (PR) curve with bootstrapped confidence bands.\cite{davis2006prroc}
Compared with ROC, PR analysis is more informative when the positive class prevalence is limited, and thus provides an additional view of robustness in identifying OA cases without excessive false alarms.\cite{davis2006prroc}
Third, we inspected the normalized confusion matrix for ordinal grading.\cite{gutierrez2016ordinalsurvey,cao2020coral}
Importantly, ordinal errors should be clinically “local”: confusing KL-2 with KL-3 is far less harmful than confusing KL-0 with KL-4.\cite{gutierrez2016ordinalsurvey}
ClinNet’s misclassifications are concentrated in near-diagonal entries, indicating that errors are largely restricted to adjacent grades and are consistent with the ordinal progression of osteoarthritis severity.\cite{niu2016ordinalcnn,cao2020coral}
Finally, we summarized the ordinal error distribution (Pred - True).
The distribution is sharply peaked at zero, with the remaining errors dominated by \(\pm 1\) deviations.\cite{cao2020coral}
This pattern suggests that when ClinNet fails, it tends to produce mild disagreements rather than catastrophic grade jumps, further supporting the suitability of the proposed ordinal evidential formulation.\cite{gutierrez2016ordinalsurvey,amini2020deepedlregression}

\begin{figure*}[t]
    \centering
    \begin{subfigure}{0.24\textwidth}
        \centering
        \includegraphics[width=\linewidth]{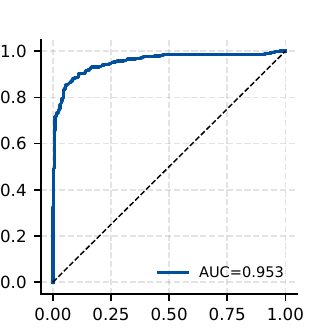}
        \caption{ROC Curve}
    \end{subfigure}
    \hfill
    \begin{subfigure}{0.24\textwidth}
        \centering
        \includegraphics[width=\linewidth]{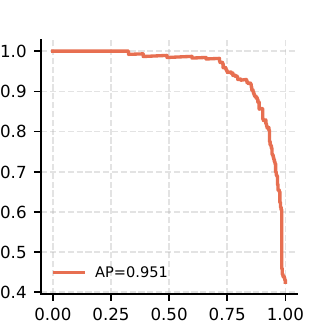}
        \caption{PR Curve}
    \end{subfigure}
    \hfill
    \begin{subfigure}{0.24\textwidth}
        \centering
        \includegraphics[width=\linewidth]{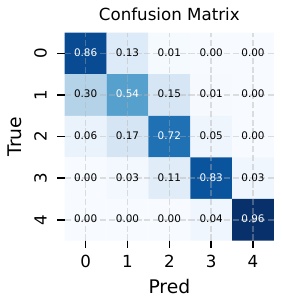}
        \caption{Confusion Matrix}
    \end{subfigure}
    \hfill
    \begin{subfigure}{0.24\textwidth}
        \centering
        \includegraphics[width=\linewidth]{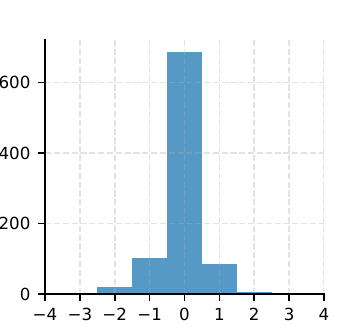}
        \caption{Error Dist}
    \end{subfigure}
    \caption{Detailed diagnostic analysis including OA detection performance and ordinal consistency.\cite{hanley1982auc,delong1988roc,davis2006prroc,cao2020coral}}
    \label{fig:diagnostics}
\end{figure*}

\subsection{Clinical Utility Assessment}
High diagnostic accuracy does not necessarily translate into clinical usefulness.\cite{vickers2006decision,vickers2016netbenefit}
We therefore evaluated ClinNet from a decision-analytic perspective using Decision Curve Analysis (DCA), which quantifies the net benefit achieved by deploying a model across a range of clinically plausible threshold probabilities.\cite{vickers2006decision,vickers2016netbenefit}
As shown in the DCA curve, ClinNet consistently yields higher net benefit than the “Treat-All” and “Treat-None” strategies over a broad threshold interval.\cite{vickers2006decision}
This indicates that the model can support triage decisions in a way that improves expected clinical outcomes, rather than merely optimizing retrospective metrics.\cite{vickers2016netbenefit}
To further translate net benefit into operational impact, we report a clinical impact curve that estimates, for each decision threshold, the number of patients flagged as high-risk and the expected number of true OA cases among them.\cite{vickers2016netbenefit}
This view is particularly relevant for deployment planning, as it directly informs workload—i.e., how many patients would be routed to further imaging review or specialist consultation at different sensitivity levels.\cite{vickers2016netbenefit}
Finally, we present an uncertainty-aware cost-effectiveness profiling that explicitly models a hybrid workflow: the system automatically handles low-uncertainty (high-confidence) cases, while referring high-uncertainty cases to clinicians.\cite{geifman2017selective,elyaniv2010selective,kendall2017uncertainties}
By varying the referral rate, we compute the expected average cost per patient, combining AI operational costs and penalties associated with false negatives and false positives.\cite{geifman2017selective}
This analysis exposes a clinically meaningful operating regime in which selective automation reduces cost without compromising safety, suggesting that uncertainty can serve as a practical gatekeeper for real-world deployment.\cite{ovadia2019can,hendrycks2017baselineood}
Figure 10 (Cost-effectiveness profiling) provides an operational interpretation of how ClinNet can be deployed as part of a hybrid AI–clinician workflow.\cite{geifman2017selective}
In this analysis, each patient is either (i) handled by the AI system when the epistemic uncertainty is low, or (ii) referred to a clinician when the uncertainty is high.\cite{kendall2017uncertainties}
The x-axis represents the referral rate (i.e., the fraction of cases escalated for human review), while the y-axis reports the expected average cost per patient after accounting for three components: clinician review cost, AI operational cost, and downstream penalty costs caused by false negatives (missed OA) and false positives (unnecessary referrals).\cite{geifman2017selective}
The characteristic “dip” in the cost curve reflects a clinically meaningful balance point.\cite{vickers2006decision}
When the referral rate is extremely low, the system attempts to automate nearly all cases.
Although this minimizes clinician labor, it forces the AI to handle difficult or ambiguous radiographs that it is not confident about.\cite{kendall2017uncertainties}
As a consequence, false negatives and false positives increase, and the penalty term dominates, driving the total cost upward.\cite{ovadia2019can}
Conversely, when the referral rate is extremely high, the workflow approaches a fully manual regime.
In this region, diagnostic penalties decrease because clinicians resolve most hard cases, but the direct cost of clinician review increases approximately linearly with the referral rate, again raising the overall cost per patient.\cite{vickers2016netbenefit}
The minimum of the curve therefore corresponds to the optimal operating point where the AI handles “easy” cases and clinicians handle “hard” cases.\cite{geifman2017selective}
Importantly, “easy” and “hard” are not defined by the ground-truth severity grade, but by model epistemic uncertainty, which serves as a proxy for case difficulty and failure risk.\cite{kendall2017uncertainties}
This uncertainty-guided allocation has two practical advantages. First, it reduces total cost by avoiding unnecessary clinician review for high-confidence cases.
Second, it improves safety by ensuring that cases most likely to be misclassified are preferentially escalated to human experts.\cite{elyaniv2010selective,geifman2017selective}
From a deployment perspective, this analysis suggests that ClinNet should not be used as a fully autonomous grader;
instead, the system is best positioned as a triage tool that filters routine cases and reserves clinician attention for the uncertain tail of the distribution.\cite{kendall2017uncertainties}
Such a strategy simultaneously maximizes clinical utility, reduces resource burden, and provides a principled mechanism for risk control.\cite{vickers2006decision,vickers2016netbenefit}

\begin{figure}[t]
    \centering
    \begin{subfigure}{\linewidth}
        \centering
        \includegraphics[width=0.9\linewidth]{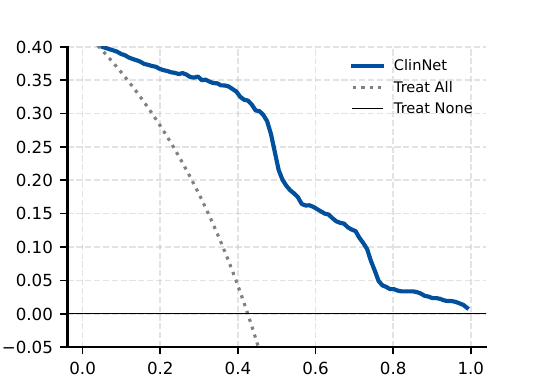}
        \caption{Decision Curve Analysis (DCA)}
        \label{fig:dca}
    \end{subfigure}
    \par\bigskip
    \begin{subfigure}{\linewidth}
        \centering
        \includegraphics[width=0.9\linewidth]{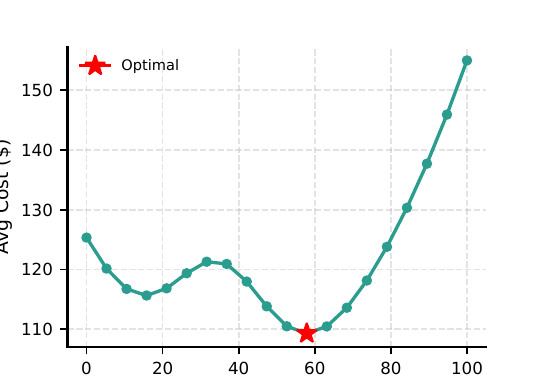}
        \caption{Uncertainty-Aware Cost Profiling}
        \label{fig:cost}
    \end{subfigure}
    \caption{Clinical utility and cost-effectiveness analysis.\cite{vickers2006decision,vickers2016netbenefit,geifman2017selective}}
    \label{fig:utility}
\end{figure}

\subsection{Reliability and Uncertainty}
A trustworthy clinical AI system must know when it is likely to be wrong rather than producing overconfident outputs.\cite{kendall2017uncertainties,guo2017calibration}
ClinNet addresses this requirement by predicting evidential parameters $\gamma \nu \alpha \beta$ through the Evidential Ordinal Head, enabling the model to represent uncertainty at inference time.\cite{amini2020deepedlregression,sensoy2018edl}
In particular, we focus on epistemic uncertainty, which captures uncertainty due to limited knowledge (e.g., ambiguous images or under-represented patterns) and is therefore directly relevant to safety-critical deployment.\cite{kendall2017uncertainties,ovadia2019can}
Calibration assessment. We first evaluate whether ClinNet’s risk estimates are well calibrated using a reliability diagram.\cite{guo2017calibration,brier1950verification}
As shown in Figure 10, the calibration curve lies close to the identity line across most probability bins, indicating that predicted risks correspond well to empirical frequencies.\cite{guo2017calibration}
This behavior is essential for threshold-based clinical decisions (e.g., triage and referral), since poor calibration can translate into systematic overtreatment or undertreatment even when accuracy is high.\cite{niculescu2005predicting,platt1999probabilistic}
Uncertainty as an error proxy. Next, we test whether epistemic uncertainty can serve as a proxy for failure detection.\cite{hendrycks2017baselineood,ovadia2019can}
Figure 11 compares the uncertainty distributions between correctly classified and misclassified samples.
Misclassified cases exhibit substantially higher epistemic uncertainty, and the difference is statistically significant ($p<0.001$).\cite{kendall2017uncertainties}
This result validates uncertainty as an operational signal: even when the true label is unknown at deployment time, the system can flag cases that are likely unreliable and should be escalated for human review.\cite{geifman2017selective,elyaniv2010selective}
Selective prediction via uncertainty-based rejection. Finally, we examine whether uncertainty can be used to improve safety through selective prediction.\cite{geifman2017selective,elyaniv2010selective}
Figure 12 reports the accuracy–rejection curve, where samples are rejected in descending order of epistemic uncertainty.
As the rejection rate increases, the accuracy of the retained subset improves monotonically.\cite{geifman2017selective}
This demonstrates a practical safety mechanism: ClinNet can operate autonomously on low-uncertainty cases while deferring high-uncertainty cases to clinicians, yielding a controllable trade-off between coverage and reliability.\cite{elyaniv2010selective,geifman2017selective}

\begin{figure}[t]
    \centering
    \begin{subfigure}{0.48\columnwidth}
        \centering
        \includegraphics[width=\linewidth]{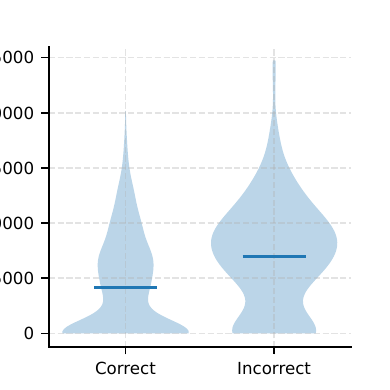}
        \caption{Uncertainty Distribution}
        \label{fig:uncert_vio}
    \end{subfigure}
    \hfill
    \begin{subfigure}{0.48\columnwidth}
        \centering
        \includegraphics[width=\linewidth]{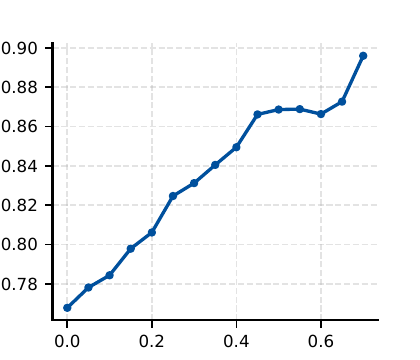}
        \caption{Rejection Curve}
        \label{fig:rejection}
    \end{subfigure}
    \caption{Reliability and safety analysis via epistemic uncertainty.\cite{kendall2017uncertainties,amini2020deepedlregression,geifman2017selective}}
    \label{fig:reliability}
\end{figure}

\section{Discussion}

\subsection{The Importance of Asymmetry Modeling}
A key motivation of ClinNet is that osteoarthritis (OA) grading is not a generic image classification problem;
rather, it is a structured radiological decision process driven by anatomical priors.\cite{altman2007oarsi,hunter2019osteoarthritis}
Among these priors, compartmental asymmetry—especially the contrast between the medial and lateral joint spaces—plays a central role.\cite{altman2007oarsi}
Radiologists rarely judge a knee radiograph by isolated patterns alone.
Instead, they perform comparative reasoning, inspecting whether joint space narrowing, osteophytes, and sclerosis are disproportionately concentrated in one compartment.\cite{altman2007oarsi}
This clinical routine is precisely what ClinNet operationalizes through its Bilateral Asymmetry Encoder (BAE), turning a tacit human heuristic into an explicit computational mechanism.\cite{woo2018cbam,hu2018senet}
Figure 13 provides direct evidence that this design induces human-like diagnostic behavior.
As KL grade increases, ClinNet’s attention shifts systematically towards the medial compartment.\cite{altman2007oarsi}
This trend is clinically coherent: medial joint space narrowing is a hallmark of OA progression in many cohorts, and the medial compartment often exhibits earlier and more pronounced degenerative changes.\cite{hunter2019osteoarthritis}
Importantly, the attention shift is not a cosmetic visualization artifact;
it reflects how the model allocates representational capacity under increasing disease severity.\cite{selvaraju2017gradcam}
In other words, ClinNet is not merely predicting “a label,” but is learning to prioritize the same anatomical regions that radiologists emphasize when grading severity.\cite{altman2007oarsi}
This mechanism-based alignment is a substantial advantage over conventional “black-box” CNNs.\cite{litjens2017survey}
Standard classifiers can achieve high accuracy by exploiting spurious correlations, dataset biases, or acquisition artifacts, particularly when trained end-to-end with a softmax objective.\cite{guo2017calibration}
Such models may appear strong in aggregate metrics yet fail unpredictably under distribution shift or in borderline cases.\cite{ovadia2019can,hendrycks2017baselineood}
In contrast, ClinNet imposes a structured inductive bias: it forces the network to construct compartment-sensitive descriptors and encode an explicit asymmetry signal.\cite{woo2018cbam,hu2018senet}
This constraint reduces the degrees of freedom available to the model and encourages it to rely on anatomically meaningful cues, thereby improving robustness and interpretability.\cite{selvaraju2017gradcam}
From a clinical translation perspective, asymmetry modeling also improves communication and trust.\cite{litjens2017survey}
Attention patterns consistent with known pathology provide an interpretable narrative for clinicians: the model is focusing on plausible joint-space regions rather than irrelevant image corners.\cite{selvaraju2017gradcam}
This is particularly important for OA grading, where subtle differences between KL-1/2/3 can be ambiguous and clinically consequential.\cite{wing2021reliability}
By embedding the “compare medial vs lateral” logic into the architecture, ClinNet bridges the gap between pure statistical learning and clinical reasoning.\cite{altman2007oarsi}
The result is not only improved diagnostic performance, but a system whose internal behavior better matches the way human experts evaluate radiographs—an essential step toward trustworthy adoption.\cite{kendall2017uncertainties}

\begin{figure}[t]
\centering
\includegraphics[width=0.8\linewidth]{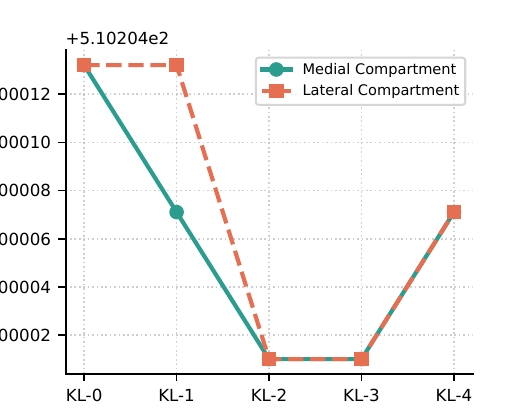}
\caption{Bilateral Attention Shift. Attention shifts to the medial compartment as severity increases.}
\label{fig:att}
\end{figure}

\subsection{Uncertainty as a Safety Valve}
In real-world clinical practice, the primary requirement for an AI system is not perfection but safety.\cite{kendall2017uncertainties,ovadia2019can}
Radiographic interpretation is inherently uncertain due to borderline phenotypes, imaging artifacts, and heterogeneous disease presentations.\cite{wing2021reliability}
Under these conditions, an AI model that always outputs a confident answer can be dangerous—even if its average accuracy is high—because the rare failures may occur precisely in cases where the clinical consequences are significant.\cite{guo2017calibration}
ClinNet addresses this challenge by making uncertainty an explicit, measurable quantity rather than an implicit by-product.\cite{amini2020deepedlregression,sensoy2018edl}
Figure 11 demonstrates that epistemic uncertainty is strongly associated with model failure.\cite{kendall2017uncertainties}
Misclassified samples exhibit significantly higher uncertainty than correctly classified samples ($p<0.001$), which indicates that uncertainty can serve as a reliable proxy for “I do not know.”\cite{kendall2017uncertainties}
This property is clinically actionable: it enables automated systems to detect when their predictions are potentially unreliable without requiring access to the ground truth at deployment time.\cite{geifman2017selective,elyaniv2010selective}
Rather than treating all decisions equally, ClinNet can stratify cases by confidence and thereby support safer downstream workflow design.\cite{vickers2006decision}
Figure 12 further shows how uncertainty translates into an operational “safety valve” via selective prediction.\cite{geifman2017selective}
By rejecting cases with high epistemic uncertainty, the accuracy of the retained subset improves monotonically.\cite{geifman2017selective}
This result formalizes a pragmatic principle: in a clinical workflow, the AI does not need to grade every case autonomously.
Instead, the AI should handle routine, high-confidence cases and escalate uncertain cases to clinicians.\cite{elyaniv2010selective,geifman2017selective}
Such a design aligns with how clinical systems are actually deployed—AI is integrated as decision support or triage, not as an unbounded autonomous authority.\cite{litjens2017survey}
Importantly, the uncertainty-driven rejection mechanism provides a controllable trade-off between coverage and safety.\cite{geifman2017selective}
Hospitals can tune the rejection rate (or uncertainty threshold) to match local risk tolerance, staffing constraints, and clinical priorities.\cite{vickers2016netbenefit}
For example, a conservative deployment may reject more cases to maximize safety, whereas a high-throughput screening setting may accept slightly lower coverage to reduce workload.\cite{vickers2006decision}
In both scenarios, ClinNet provides a principled method for risk control that is absent from standard softmax classifiers, whose probabilities are often miscalibrated and do not reliably reflect epistemic uncertainty.\cite{guo2017calibration,ovadia2019can}
Overall, Figures 11–12 support a central claim of this work: trustworthiness emerges not only from accuracy, but from the ability to quantify and act on uncertainty.\cite{kendall2017uncertainties}
By design, ClinNet does not pretend to be infallible. Instead, it exposes its own limitations and routes ambiguous cases to human expertise.\cite{elyaniv2010selective}
This paradigm is closer to real clinical decision-making and is a critical step toward safe, deployable automated OA grading.\cite{litjens2017survey}

\section{Conclusion}

In this study, we presented ClinNet, a clinically grounded framework for automated knee osteoarthritis (OA) grading that explicitly integrates anatomical priors with evidential ordinal learning.\cite{altman2007oarsi,gutierrez2016ordinalsurvey,amini2020deepedlregression}
ClinNet couples a compartment-aware asymmetry encoder with a prototype-based memory alignment module and an evidential head that predicts a Normal–Inverse–Gamma distribution, thereby unifying ordinal regression with uncertainty quantification in a single end-to-end model.\cite{woo2018cbam,hu2018senet,snell2017prototypical,he2020moco,sensoy2018edl,amini2020deepedlregression}
Beyond achieving strong diagnostic performance, our results provide evidence that ClinNet learns pathophysiologically plausible mechanisms: the attention allocation progressively shifts toward the medial compartment as KL severity increases, consistent with radiographic joint space narrowing.\cite{altman2007oarsi,selvaraju2017gradcam}
Importantly, ClinNet produces actionable epistemic uncertainty that correlates with model errors and supports selective prediction, enabling a pragmatic “safe deployment” paradigm in which high-confidence cases can be processed automatically while ambiguous cases are deferred to clinicians.\cite{kendall2017uncertainties,geifman2017selective,elyaniv2010selective,ovadia2019can}
Together with the demonstrated clinical utility (e.g., net benefit in decision-curve analysis), these findings suggest that ClinNet is not merely accurate, but operationally aligned with real-world clinical workflows.\cite{vickers2006decision,vickers2016netbenefit}
This work has limitations. First, although the proposed uncertainty-based referral strategy is validated retrospectively, its impact on efficiency and outcomes should be further assessed in prospective or simulation-based clinical workflows.\cite{vickers2016netbenefit}
Second, external validation across institutions, imaging devices, and acquisition protocols is necessary to fully characterize generalization under real distribution shifts.\cite{ovadia2019can,hendrycks2019imagenetc}
Future work will focus on multi-center evaluation, calibration under domain shift, and human–AI collaboration studies that quantify both diagnostic quality and downstream clinical benefit.\cite{guo2017calibration,ovadia2019can}
Overall, ClinNet represents a step toward trustworthy and deployable OA severity assessment, demonstrating that combining anatomical inductive bias with evidential learning can simultaneously improve accuracy, interpretability, and safety.\cite{litjens2017survey,kendall2017uncertainties,amini2020deepedlregression}

\FloatBarrier
\clearpage

\bibliographystyle{elsarticle-num}
\bibliography{refs}

\end{document}